# SERKET: An Architecture for Connecting Stochastic Models to Realize a Large-Scale Cognitive Model

**Tomoaki Nakamura**[1,*], **Takayuki Nagai**[1], **Tadahiro Taniguchi**[2]

[1]*Department of Mechanical Engineering and Intelligent Systems, The University of Electro-Communications, Tokyo, Japan*
[2]*College of Information Science and Engineering, Ritsumeikan University, Shiga, Japan*

Correspondence*:
Tomoaki Nakamura
tnakmaura@uec.ac.jp

## ABSTRACT

To realize human-like robot intelligence, a large-scale cognitive architecture is required for robots to understand the environment through a variety of sensors with which they are equipped. In this paper, we propose a novel framework named Serket that enables the construction of a large-scale generative model and its inference easily by connecting sub-modules to allow the robots to acquire various capabilities through interaction with their environments and others. We consider that large-scale cognitive models can be constructed by connecting smaller fundamental models hierarchically while maintaining their programmatic independence. Moreover, connected modules are dependent on each other, and parameters are required to be optimized as a whole. Conventionally, the equations for parameter estimation have to be derived and implemented depending on the models. However, it becomes harder to derive and implement those of a larger scale model. To solve these problems, in this paper, we propose a method for parameter estimation by communicating the minimal parameters between various modules while maintaining their programmatic independence. Therefore, Serket makes it easy to construct large-scale models and estimate their parameters via the connection of modules. Experimental results demonstrated that the model can be constructed by connecting modules, the parameters can be optimized as a whole, and they are comparable with the original models that we have proposed.

## 1 INTRODUCTION

To realize human-like robot intelligence, a large-scale cognitive architecture is required for robots to understand the environment through a variety of sensors with which they are equipped. In this paper, we propose a novel framework that enables the construction of a large-scale generative model and its inference easily by connecting sub-modules in order that the robots acquire various capabilities through interactions with the environment and others. We consider that it is important for robots to understand the real world by learning from environment and others, and have proposed a method that enables robots to acquire concepts and language (Attamimi et al., 2016; Nakamura et al., 2014; Nishihara et al., 2017; Taniguchi et al., 2017) based on the clustering of multimodal information obtained by the robots. These proposed models are based on Bayesian models with a complicated structure, and, we derived and implemented the equations of parameter estimation. If we realize a model that enables robots to learn more complicated capabilities, we have to construct a more complicated model, and derive and implement equations for parameter estimation. Therefore, it is considered to be difficult to construct higher level cognitive models by leveraging this approach. Alternatively, these models can be interpreted as a composition of more fundamental Bayesian models. In this paper, we consider developing a large-scale cognitive model by connecting the Bayesian models and propose an architecture named Serket (Symbol Emergence in Robotics tool KIT) that enables the construction of such models more easily.

In the field of cognitive science, cognitive architectures (Laird, 2008; Anderson, 2009) have been proposed to implement human cognitive mechanisms by describing humans perception, judgement and diction making. However, complex machine learning algorithms have not yet been introduced, and it is difficult to implement the models that we have proposed. However, Serket makes it possible to implement more complex models by connecting modules.

As one approach to develop a large-scale cognitive model, the probabilistic programming languages (PPLs), which make it easy to construct Bayesian models, have been proposed (Carpenter et al., 2016; Patil et al., 2010; Tran et al., 2016; Wood et al., 2014; Goodman et al., 2012). The advantages are that PPLs can construct Bayesian models by defining dependencies between the random variables, and the parameters are automatically estimated without deriving the equations for it. By using PPLs, it is easy to construct relatively small-scale models such as a Gaussian mixture model and a latent Dirichlet allocation, and it is still difficult to model multimodal sensory information such as images and speech obtained by the robots. Due to this, we implemented models for concept and language acquisition, which are relatively large-scale models, as standalone models without PPLs. However, we consider the approach where an entire model is implemented by itself has limitations if it is constructed as a large-scale model.

We consider that large-scale cognitive models can be constructed by connecting smaller fundamental models hierarchically, and, actually, our proposed models have such structure. In the proposed novel architecture Serket, the large-scale models are constructed by hierarchically connecting smaller scale Bayesian models (hereinafter one of these is called a *module*) while maintaining their programmatic independence. Moreover, connected modules are dependent upon each other, and parameters are required to be optimized as a whole. When the models are constructed by themselves, the equations for parameter estimation have to be derived and implemented depending on the models. However, in this paper, we propose a method for parameter estimation by communicating the minimal parameters between various modules while maintaining their programmatic independence. Therefore, Serket makes it easy to construct large-scale models and estimate their parameters by connecting modules.

In this paper, we propose the Serket framework and implement the models that we have proposed by leveraging this framework. Experimental results demonstrated that the model can be constructed by connecting modules, the parameters can be optimized as a whole, and they are comparable with original models that we have proposed.

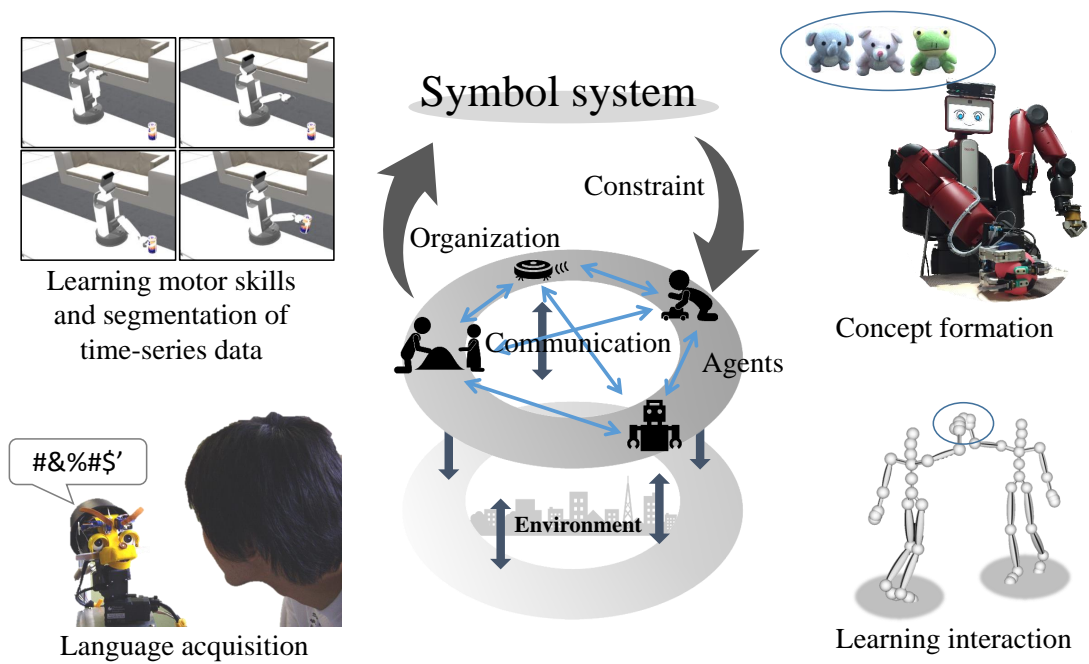

**Figure 1.** Symbol emergence system.

## 2 BACKGROUND

### 2.1 Symbol Emergence in Robotics

Recently, it has been said that artificial intelligence is superior to human intelligence in supervised learning as typified by deep learning as far as certain specific tasks (He et al., 2015; Silver et al., 2017). However, we believe that it is difficult to realize human-like intelligence only via supervised learning, because all supervised labels cannot be obtained for all the sensory information of robots. To this end, we believe that it is also important for robots to understand the real environment by structuring their own sensory information in an unsupervised manner, and we consider such a learning process as a symbol emergence system (Taniguchi et al., 2016a).

The symbol emergence system is based on the genetic epistemology proposed by Piaget (Piaget and Duckworth, 1970). In genetic epistemology, humans organize symbol systems in a bottom-up manner through interaction with the environment. Fig. 1 depicts an overview of the symbol emergence system. The symbols are self-organized from sensory information that is obtained through interactions with the environment. However, it can be difficult for the robots to interact with others using symbols learned only in a bottom-up manner because the sensory information cannot be shared with others directly and the meaning of symbols differs depending on individuals. To interact with others, the symbols are shared among individuals and the meanings of the symbols are transformed through such interactions. In fact, these symbols affect the individual organization of symbols in a top-down manner. Thus, in the symbol emergence system, the symbols emerge through loops of top-down and the bottom-up effects. As symbols, we consider not only linguistic symbols but also self-organized concepts that represent various types of knowledge about the environment and others. Therefore, SER covers many research topics such as concept formation (Nakamura et al., 2007), language acquisition (Nishihara et al., 2017; Taniguchi et al., 2017, 2016b), learning of interactions (Taniguchi et al., 2010), learning of body scheme (Mimura et al., 2017), and learning of motor skills and segmentation of time-series data (Taniguchi et al., 2011; Nakamura et al., 2016).

We have proposed models that enable robots to acquire concepts and language by considering its learning process as a symbol emergence system. The robots form concepts in a bottom-up manner, and acquire word meanings by connecting words and concepts. Simultaneously, words are shared with others, and their meanings are changed through communication with others. Therefore, such words affect concept formation in a top-down manner, and concepts are changed. Thus, we have considered that robots can acquire concepts and word meanings through loops of bottom-up and top-down effects.

## 2.2 Existing Cognitive Architecture

There have been many efforts to develop intelligent systems. In the fields of cognitive science, cognitive architectures (Laird, 2008; Anderson, 2009) have been proposed to implement humans cognitive mechanisms by describing humans perception, judgment and decision making. However, as mentioned earlier, we consider that it is important to consider how to model the multimodal sensory information obtained by robots, however this is still difficult to achieve with these cognitive architectures.

The frameworks of deep neural networks (DNN) such as TensorFlow (Abadi et al., 2016), Keras (Chollet et al., 2015), and Chainer (Tokui et al., 2015) have been developed. These frameworks make it possible to construct deep neural network models and estimate parameters easily. These frameworks are considered to be one of the factors explaining why DNNs have been widely used for several years.

Alternatively, probabilistic programming languages (PPLs), which make it easy to construct Bayesian models, have been proposed (Carpenter et al., 2016; Patil et al., 2010; Tran et al., 2016; Wood et al., 2014; Goodman et al., 2012). The advantages of PPLs are that they can construct Bayesian models by defining dependencies between random variables, and parameters are automatically estimated without deriving the equations for them. By using PPLs, relatively small-scale models such as the Gaussian mixture model and latent Dirichlet allocation can be constructed easily. However, it is still difficult to model the multimodal sensory information such as images and speech obtained by the robots. We believe a framework is required to construct a large-scale probabilistic generative model to model the multimodal information of the robot.

## 2.3 Cognitive Architecture Based on Probabilistic Generative Model

We believe cognitive models are models that make it possible to predict an output $Y$ against an input $X$. For example, as shown in Fig. 2, an object label $Y$ is predicted from a sensor input $X$ via object recognition, and it is via an understanding of word meanings that the semantic contents $Y$ are predicted from a speech signal $X$. In other words, the problem can be defined as: how to model $P(Y|X)$, where the prediction is realized by $\mathrm{argmax}_Y P(Y|X)$. Deep neural networks model relationships between an input $X$ and output $Y$ directly by an end-to-end approach (Fig. 2(b)). Alternatively, we consider developing these cognitive models by leveraging Bayesian models where $X$ and $Y$ are treated as random variables and the relationships between them are represented by a latent variable $Z$ (Fig. 2(a)). Therefore, in the Bayesian models, the prediction of the output $Y$ from the input $X$ is computed as follows:

$$P(Y|X) \propto P(Y,X) \tag{1}$$
$$= \int_Z P(Y|Z)P(X|Z)P(Z)dZ. \tag{2}$$

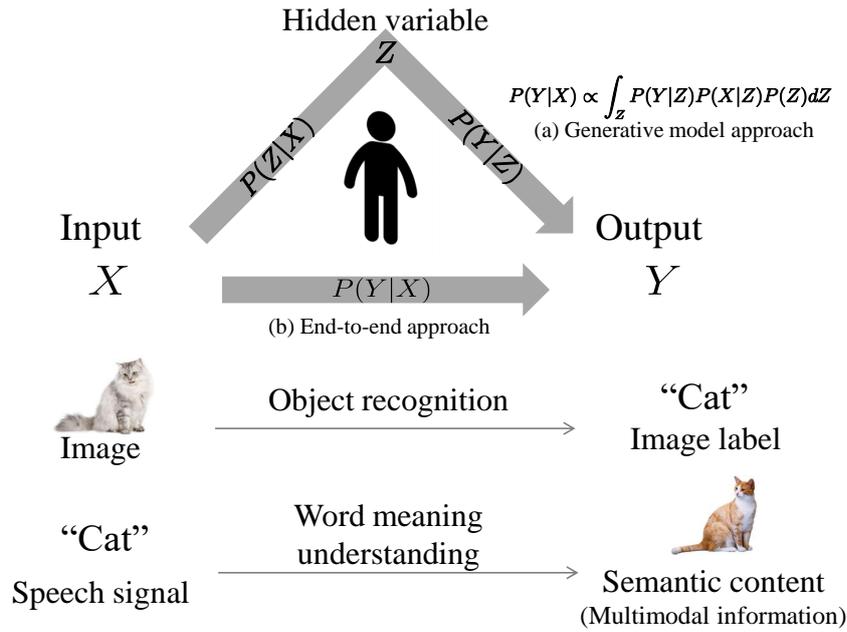

**Figure 2.** Overview of cognitive model by (a) probabilistic generative model and (b) end-to-end learning.

This is multimodal latent Dirichlet allocation that we have proposed in (Nakamura et al., 2012), the details of which are described in the appendix of this paper. However, this is based on an important assumption that the observed variables $X$ and $Y$ are conditionally independent against the latent variable $Z$. Here, we consider models where assumptions about multiple observations are made without distinguishing between input and output. Fig. 3(a) displays the generalized model. In this generalized model, the right side of Eq. (1) corresponds to the following equation, and a part of the observations can be predicted from other observations:

$$P(\boldsymbol{o}_1, \boldsymbol{o}_2, \cdots) = \int_z P(z) \Pi_n P(\boldsymbol{o}_n|z) dz. \tag{3}$$

However, as mentioned earlier, it is assumed that all observations $\boldsymbol{o}_1, \boldsymbol{o}_2, \cdots$ are conditionally independent against $Z$. Considering the modeling of sensor information as observations $\boldsymbol{o}_1, \boldsymbol{o}_2, \cdots$, it is difficult for all observations to satisfy this assumption. To overcome this problem, this model is hierarchized until all observations are independent as shown in Fig. 3(b). In this hierarchized model, $o_{*,*}$ are observations and $z_{*,*}$ are latent variables, and the right side of Eq. (1) corresponds to a following equation:

$$P(\boldsymbol{O}|z_{M,1}) = \prod_m^M \prod_n^{\bar{N}_m} \int_{z_{m,n}} P(z_{m,n}) \prod_i^{N_m} P(\boldsymbol{o}_{m,n,i}|z_{m,n}) \prod_{n'}^{\bar{N}_{m-1}} P(z_{m-1,n'}|z_{m,n}) dz_{m,n}, \tag{4}$$

where $\boldsymbol{O}$ is the set of all observation, $M$ is the number of the hierarchy, and $N_m$ and $\bar{N}_m$ denote the number of observations and latent variables in the $m$-th hierarchy. In this model, it is not difficult to analytically derive equations to estimate the parameters if the number of the hierarchy is not large. However, it is more difficult to derive them if the number of the hierarchy increases. To estimate the parameters of the hierarchical model, we propose Serket, which is an architecture that renders it possible to approximately estimate the parameters by dividing them into even hierarchies.

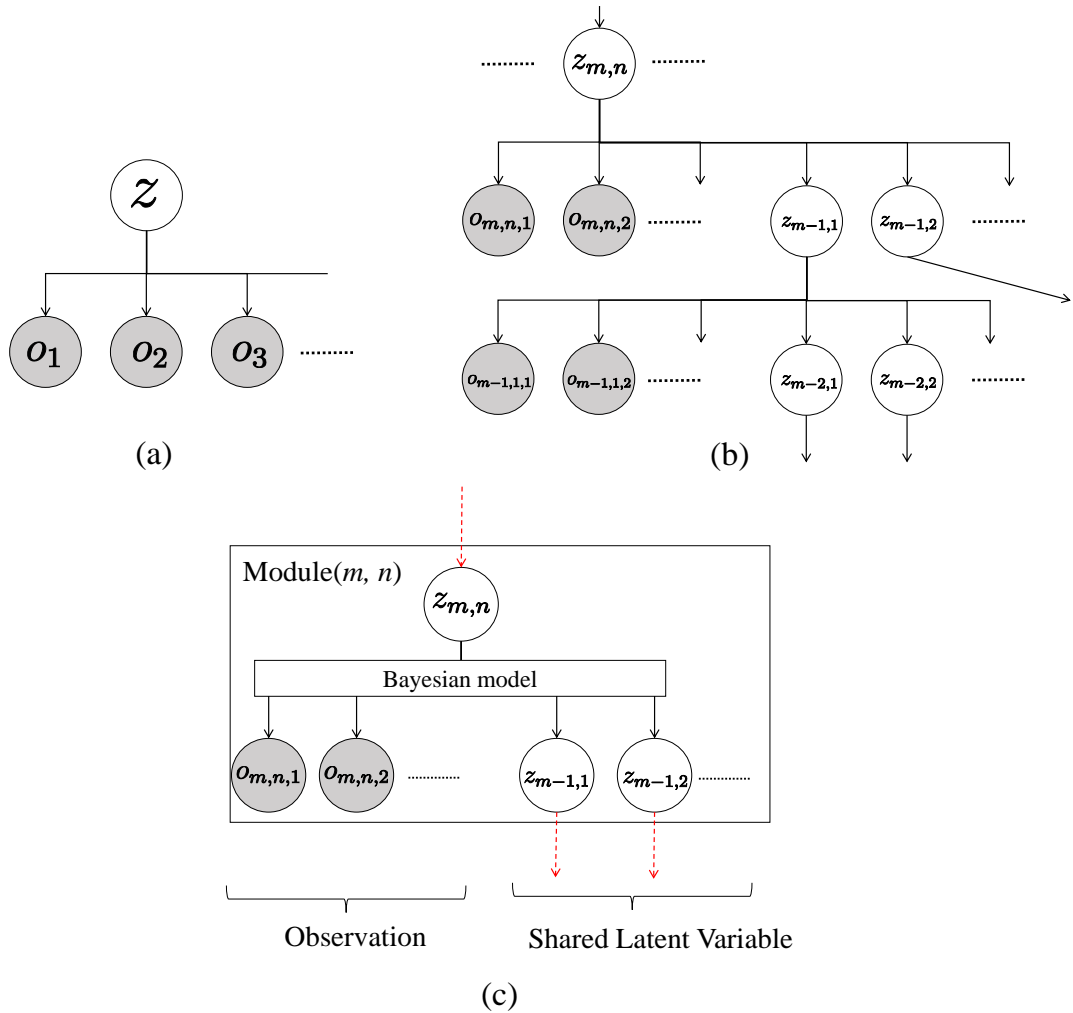

**Figure 3.** Generalized hierarchical cognitive model: (a) Single layer model, (b) multi-layered model by hierarchicalization of single layer models, and (c) the generalized form of a module in Serket.

## 2.4 Cognitive Models

In the past, studies on how the relationships between multimodal information are modeled have been conducted (Roy and Pentland, 2002; Wermter et al., 2004; Ridge et al., 2010; Ogata et al., 2010; Lallee and Dominey, 2013; Zhang et al., 2017). Neural networks were used in these studies which allowed inferences to be made based on observed information possible by learning multimodal information such as words, visual information and a robot's motions. As mentioned earlier, these are considered some of the examples of the cognitive models that we defined.

There are also studies in which manifold learning is used for modeling a robot's multimodal information (Mangin and Oudeyer, 2013; Mangin et al., 2015; Chen and Filliat, 2015; Yürüten et al., 2013). In these studies, they used manifold learning such as non-negative matrix factorization (NMF), and multimodal information, which is an observation of the model represented by low dimensional hidden parameters. We consider this as another approach to constructing cognitive models, in which the information is inferred through hidden parameters.

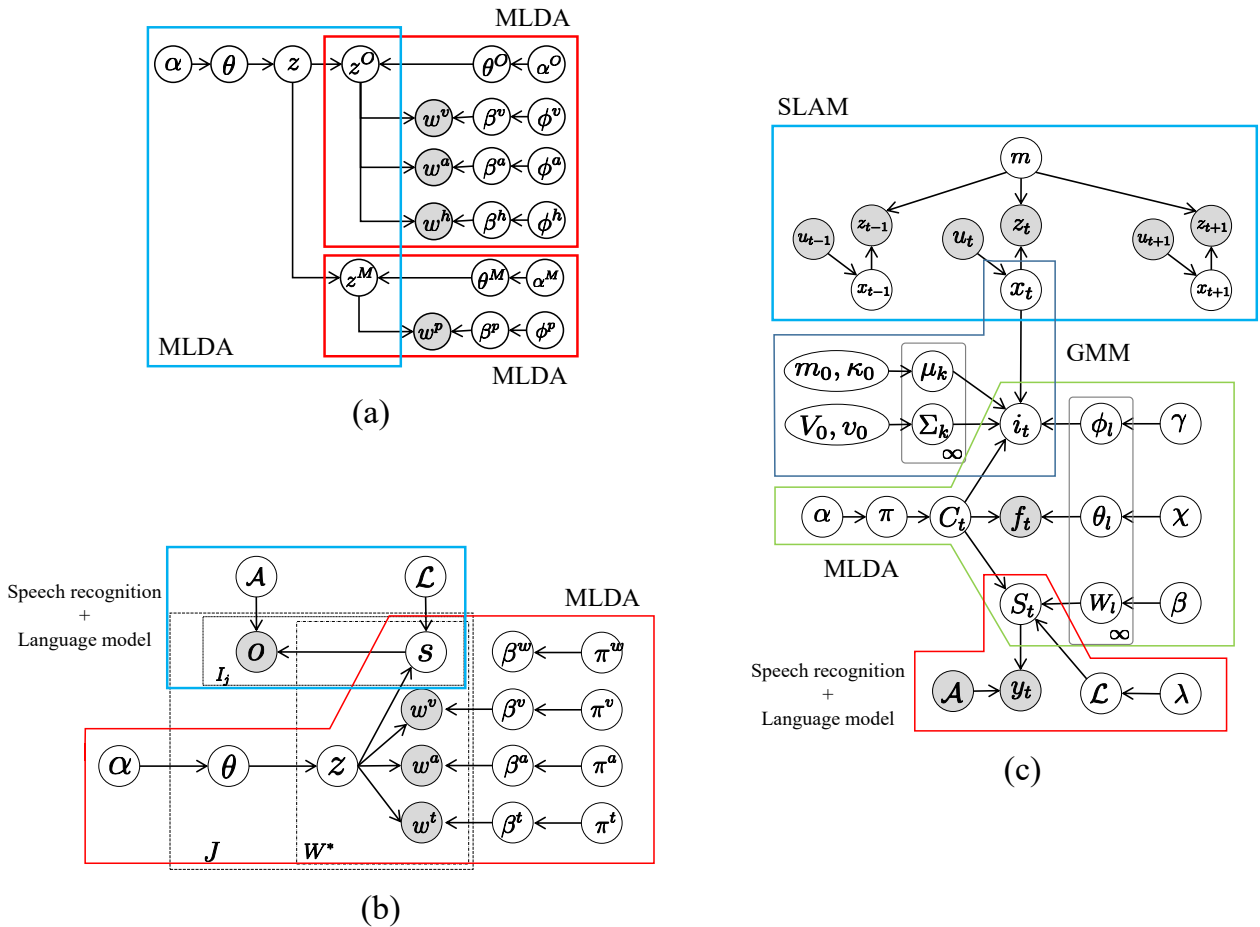

**Figure 4.** Graphical models for concept formation: (a) model for hierarchical concept(Attamimi et al., 2016), (b) model for object concept and language acquisition(Nakamura et al., 2014; Nishihara et al., 2017), and (c) model for location concept and language acquisition (Taniguchi et al., 2017).

Recently, deep neural networks have remarkably progressed in many areas such as object recognition (He et al., 2015), object detection (Redmon et al., 2016), speech recognition (Amodei et al., 2016), sentence generation (Vinyals et al., 2015), machine translation (Sutskever et al., 2014), and visual question answering (Wu et al., 2016). In these studies, end-to-end learning is used, and this makes it possible to infer information from other information. Therefore, these are also considered part of the cognitive model we defined in this paper. However, as we mentioned in Sec. 2.1, we believe it is important for robots to understand the real environment by structuring their own sensory information in an unsupervised manner.

To develop a cognitive model where the robots learn autonomously, our group proposed several models for concept formation (Nakamura et al., 2007), language acquisition (Nishihara et al., 2017; Taniguchi et al., 2017, 2016b), learning of interactions (Taniguchi et al., 2010), learning of body scheme (Mimura et al., 2017), and learning motor skills and segmentation of time-series data (Taniguchi et al., 2011;

Nakamura et al., 2016). Although all of these are targets of Serket, we in particular deal with concept formation in this paper. We define concepts as categories into which the sensory information is classified, and have proposed various concept models. These are implementations of the aforementioned hierarchical model. Fig. 4(a) displays one of our proposed models, and this is the most simple form of the hierarchical model where $z^O$ and $z^M$ denote an object and a motion concept respectively and their relationships are represented by $z$ (Attamimi et al., 2016). Therefore, in this model, $z$ represents objects and possible motions against them, which are considered as their usage, and observations become conditionally independent by introducing latent variables $z^O$ and $z^M$.

In these Bayesian models, the latent variables are shown as the white nodes $z, z^O$ and $z^M$ in Fig. 4(a) can be learned from the observations shown as gray noes in an unsupervised manner. Moreover, these latent variables are not determined independently but optimized as a whole by depending each other. Although it seems that this model has a complicated structure and it is difficult to estimate the parameters and determine the latent variables, this model can be divided into smaller components each of which is multimodal latent Dirichlet allocation. The models shown in Fig. 4(b) and (c) can also be divided into smaller components, although they have a complicated structure. Similar to these models, it is possible to develop larger models by combining smaller models as modules. In this paper, we propose a novel architecture symbol emergence in robotics tool kit (Serket) to develop larger models by combining modules.

In the proposed architecture, the parameters of each module are not learned independently but learned based on their dependence on each other. To implement such learning, it is important to share latent variables between modules. For example, $z^O$ and $z^M$ are respectively shared between two MLDAs in the model shown in Fig. 4(a). The shared latent variables are not determined independently but determined depending on each other. The proposed Serket makes it possible for each module to maintain their independence as a program as well as be learned as a whole through the shared latent variables.

## 3 SERKET

### 3.1 Composing Cognitive Sub-Modules

Fig. 3(c) displays a generalized form of the module assumed in Serket. Each module has multiple shared latent variables $z_{m-1,*}$ and observations $\boldsymbol{o}_{m,n,*}$, which are assumed to be generated from latent variable $z_{m,n}$ of a higher level. However, modules which have no shared latent variable or no observations are included in this generalized model. Moreover, the modules can have any internal structure as long as they have shared latent, observation and higher level latent variables. Based on this module, a larger model can be constructed by connecting latent variables of module$(m-1,1)$, module$(m-1,2)$, $\cdots$ recursively. In the Serket architecture, each module has to satisfy the following requirements:

1. In each module that has shared latent variables, the following probability that the latent variables are generated can be computed:

$$P(z_{m-1,i}|z_{m,n}, \boldsymbol{o}_{m,n,1}, \boldsymbol{o}_{m,n,2}, \cdots, \boldsymbol{z}_{m-1}). \tag{5}$$

2. The module can send the following probability by leveraging one of the two methods explained in the next section:

$$P(z_{m-1,i}|z_{m,n}, \boldsymbol{o}_{m,n,1}, \boldsymbol{o}_{m,n,2}, \cdots, \boldsymbol{z}_{m-1}). \tag{6}$$

3. The module can determine $z_{m,n}$ by using the following probability sent from module$(m+1, j)$ by one out of the two methods explained in the next section:

$$P(z_{m,n}|z_{m+1,j}, \boldsymbol{o}_{m+1,j,1}, \boldsymbol{o}_{m+1,j,2}, \cdots, \boldsymbol{z}_m) \tag{7}$$

4. Terminal modules have no shared latent variables and only have observations.

In Serket, the modules affecting each other and the shared latent variables are determined by their communication with each other. Methods to determine the latent variables are classified into two types depending on their nature. One is the case that they are discrete and finite, and another is the case that they are continuous or infinite.

## 3.2 Inference of composed models

### 3.2.1 Message Passing Approach

First, we consider the case that the latent variables are discrete and finite. For example, in the model shown in Fig. 4(a), the shared latent variable $z^O$ is generated from a multinomial distribution, which is represented by finite dimensional parameters. Here, we consider the estimation of the latent variables according to the simplified model shown in Fig. 5(a). In module 2, the shared latent variable $z_1$ is generated from $z_2$, and, in module 1, the observation $o$ is generated from $z_1$. The latent variable $z_1$ is shared in module 1 and module 2, and determined by the effect on these two modules as follows:

$$z_1 \sim P(z_1|\boldsymbol{o}, z_2) \tag{8}$$
$$\propto P(z_1|\boldsymbol{o})P(z_1|z_2). \tag{9}$$

In this equation, $P(\boldsymbol{o}|z_1)$ can be computed in module 1, and $P(z_1|z_2)$ can be computed in module 2. We assumed that the latent variable is discrete and finite, and $P(z_1|z_2)$ is a multinomial distribution that can be represented by a finite dimensional parameter whose dimension ranges from the number of elements of $z_1$. Therefore, $P(z_1|z_2)$ can be sent from module 2 to module 1. Moreover, $P(z_1|z_2)$ can be learned in module 2 by using $P(z_1|\boldsymbol{o})$ sent from module 1, which is also a multinomial distribution. The parameters of these distributions can be easily sent and received, and the shared latent variable can be determined by the following procedure.

1. In module 1, $P(z_1|\boldsymbol{o})$ is computed.
2. $P(z_1|\boldsymbol{o})$ is sent to module 2.
3. In module 2, the probability distribution $P(z_1|z_2)$, which represents the relationships between $z_1$ and $z_2$, is estimated using $P(z_1|\boldsymbol{o})$
4. $P(z_1|z_2)$ is sent to module 1.
5. In module 1, the latent variable $z_1$ is estimated using Eq. (9), and the parameters of $P(\boldsymbol{o}|z_1)$ are updated.

Thus, in the case that the latent variable is infinite and discrete, the modules are learned by sending and receiving the parameters of a multinomial distribution of $z_1$. In this paper, we call this approach a message passing (MP) approach because the parameters of the model can be optimized through communicating the message.

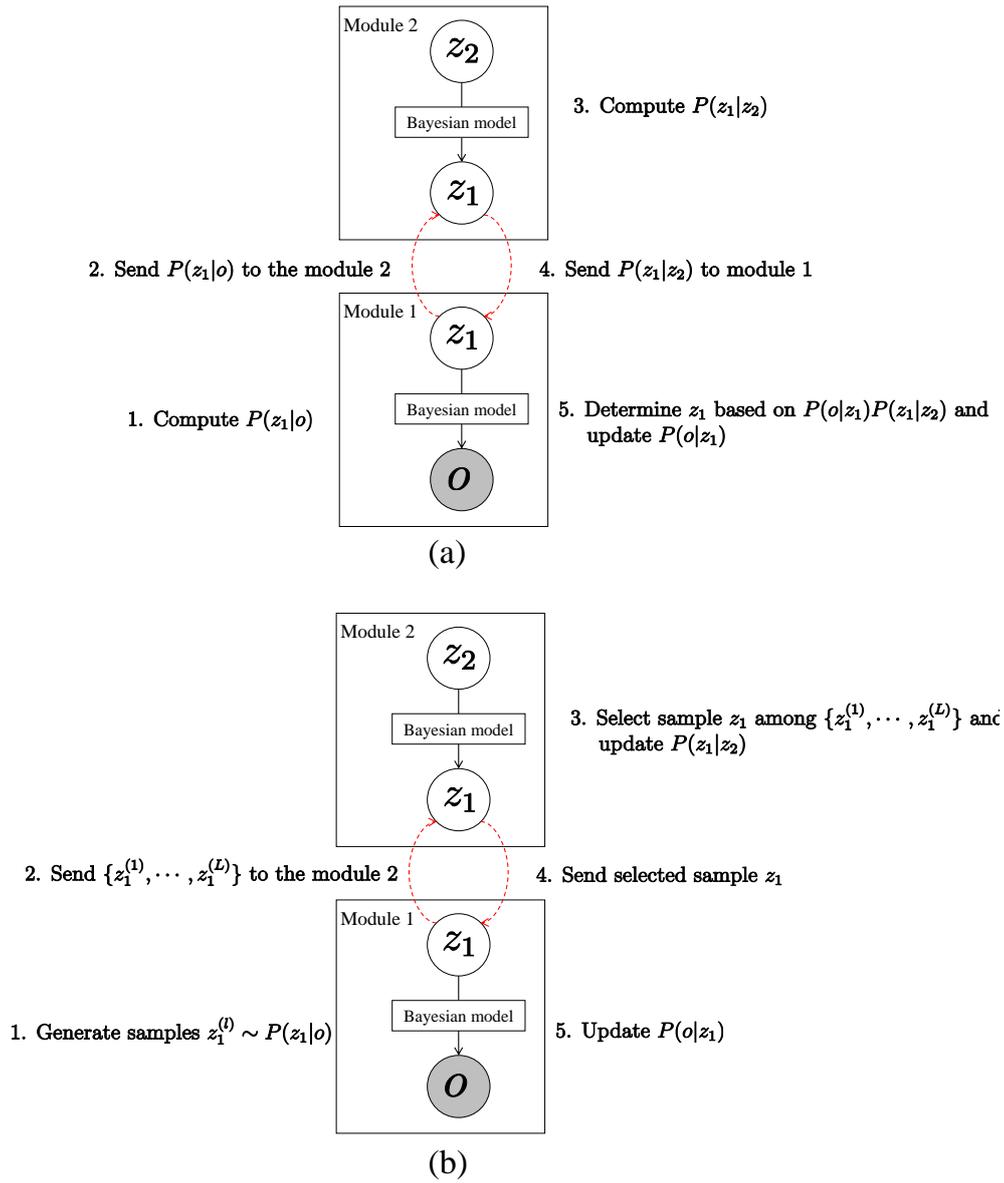

**Figure 5.** Connecting two modules by (a) MP approach and (b) SIR approach.

### 3.2.2 Sampling Importance Resampling Approach

In the previous section, the latent variable was able to be determined by communicating the parameters of the multinomial distributions if the latent variables are discrete and finite. Otherwise it can be difficult to communicate the parameters. For example, the number of the parameters becomes infinite if possible values of the latent variables are infinite patterns. Also, in the case that the form of a probability distribution is complicated, it is difficult to represent it by a small number of parameters. In these cases, the parameters of the model are learned by approximation using sampling importance resampling (SIR). We also consider the parameter estimation by using a simplified model shown in Fig. 5(b). Here, the latent variable $z_1$ is shared, and its possible value is either an infinite patterns or continuous. Similar to the previous section, the latent variable is determined if the following equation can be computed:

$$z_1 \sim P(z_1|\boldsymbol{o}, z_2) \tag{10}$$
$$\propto P(z_1|\boldsymbol{o})P(z_1|z_2). \tag{11}$$

However, in the case that the value of $z_1$ is an infinite or continuous, module 2 cannot send $P(z_1|z_2)$ to module 1. Therefore, first, $P(z_1|\boldsymbol{o})$ is approximated by $L$ samples $\{z^{(l)} : l = 1, \cdots, L\}$:

$$z_1^{(l)} \sim P(z_1|\boldsymbol{o}). \tag{12}$$

This approximation is equivalent to approximating $P(z_1|o)$ by the following $\tilde{P}(z_1|\boldsymbol{o})$:

$$P(z_1|\boldsymbol{o}) \approx \tilde{P}(z_1|\boldsymbol{o}) = \frac{1}{L}\sum_{l}^{L} \delta(z_1, z_1^{(l)}), \tag{13}$$

where $\delta(a, b)$ represents a delta function, which is 1 if $a = b$, 0 otherwise. The generated samples are sent from module 1 to module 2, and a latent variable is selected among them based on $P(z_1|z_2)$:

$$z_1 \sim P(z_1 \in \{z_1^{(1)}, \cdots, z_1^{(L)}\}|z_2). \tag{14}$$

This procedure is equivalent to sampling from the following distribution that is an approximation of Eq. (11):

$$z_1 \sim P(z_1|z_2)\tilde{P}(z_1|\boldsymbol{o}). \tag{15}$$

Thus, the parameters of each module can be updated by the determined latent variables.

### 3.2.3 Other Approaches

We previously presented two methods, however, these are not the only methods for parameter estimation. There are other applicable methods to estimate parameters. For example, one of the applicable methods is the Metropolis-Hastings (MH) approach. In the MH approach, samples are generated from a proposal distribution $Q(z|z^*)$ where $z^*$ and $z$ represent a current value and generated value of latent variables respectively. Then, they are accepted according to the acceptance probability $A(z, z^*)$:

$$A(z, z^*) = \min(1, \alpha) \tag{16}$$
$$\alpha = \frac{P(z^*)Q(z|z^*)}{P(z)Q(z^*|z)}, \tag{17}$$

where $P(z)$ represents the target distribution from which the samples are generated.

The parameters of the model shown in Fig. 5 can be estimated by considering $P(z_1|\boldsymbol{o})$ as the proposal distribution and $P(z_1|z_2, \boldsymbol{o})$ as the target distribution. $P(z_1|z_2, \boldsymbol{o})$ can be transformed to following form:

$$P(z_1|z_2, \boldsymbol{o}) \propto P(z_1|\boldsymbol{o})P(z_1|z_2)P(z_2). \tag{18}$$

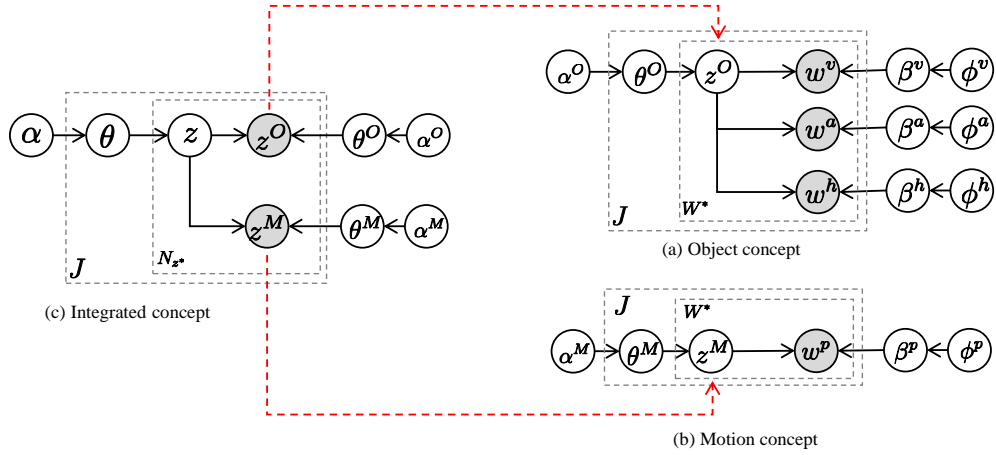

**Figure 6.** Implementation of mMLDA by connecting three MLDAs. The dashed arrows denote the conditional dependencies represented by Serket.

Therefore, $\alpha$ in Eq. (16) becomes as follows:

$$\alpha = \frac{P(z^*)Q(z|z^*)}{P(z)Q(z^*|z)} = \frac{P(z_1^*|z_2, \boldsymbol{o})}{P(z_1|z_2, \boldsymbol{o})} \cdot \frac{P(z_1|\boldsymbol{o})}{P(z_1^*|\boldsymbol{o})} \tag{19}$$

$$= \frac{P(z_1^*|\boldsymbol{o})P(z_1^*|z_2)P(z_2)}{P(z_1|\boldsymbol{o})P(z_1|z_2)P(z_2)} \cdot \frac{P(z_1|\boldsymbol{o})}{P(z_1^*|\boldsymbol{o})} = \frac{P(z_1^*|z_2)}{P(z_1|z_2)}, \tag{20}$$

Hence, the proposal distribution $P(z_1|\boldsymbol{o})$ can be computed in module 1, and the acceptance distribution can be computed in module 2. By using this approach, the parameters can be estimated while maintaining programmatic independences. The proposed value is sent to module 2, and module 2 determines if it is accepted or not. Then, the parameters are updated according to the accepted values.

Thus, various approaches can be utilized for parameter estimation, and it should be discussed which methods are most suitable. However, we leave this for a future discussion because of a limitation of space.

## 4 EXAMPLE 1: MULTILAYERED MLDA

### 4.1 Implementation Based on Serket

First, we show that a more complicated model can be constructed by combining simpler models based on Serket. The mMLDA shown in Fig. 4(a) can be constructed by using an MP approach. This model can be divided into to three MLDAs. In the lower level MLDAs, object categories $z^O$ can be formed from multimodal information $\boldsymbol{w}^v, \boldsymbol{w}^a, \boldsymbol{w}^h$ obtained from the objects, and motion categories $z^M$ can be formed from joint angles obtained by observing a human's motion. Details of the information are explained in the Appendix. Moreover, in the higher level MLDA, integrated categories $z$ that represent the relationships between objects and motions can be formed by considering $z^O$ and $z^M$ as an observation. In this model, latent variables $z^O$ and $z^M$ are shared, and, therefore, the whole model parameters are optimized in a mutually affecting manner. Fig. 6 shows the mMLDA represented by three MLDAs.

First, in the two MLDAs shown in Fig. 6(a) and (b), the probabilities $P(z_j^O|\boldsymbol{w}_j^v, \boldsymbol{w}_j^a, \boldsymbol{w}_j^h)$ and $P(z_j^M|\boldsymbol{w}_j^p)$ that the object and motion category of the $j$-th observation become $z_j^O$ and $z_j^M$ respectively and can be computed by using Gibbs sampling. These probabilities are represented by finite and discrete parameters,

which can be sent to the integrated concept model shown in Fig. 6(c). In the integrated concept model, $\hat{z}_j^O$ and $\hat{z}_j^M$ can be treated as observed variables by using these probabilities.

$$\hat{z}_{jn}^O \sim P(z_j^O|\boldsymbol{w}_j^v, \boldsymbol{w}_j^a, \boldsymbol{w}_j^h), \tag{21}$$

$$\hat{z}_{jn}^M \sim P(z_j^M|\boldsymbol{w}_j^p). \tag{22}$$

Thus, in the integrated concept model, category $z$ can be formed in an unsupervised manner. Next, the values of the shared latent variables are inferred stochastically using a learned integrated concept model:

$$P(z^O|\hat{\boldsymbol{z}}_j^M, \hat{\boldsymbol{z}}_j^O) = \sum_z P(z^O|z)P(z|\hat{\boldsymbol{z}}_j^m, \hat{\boldsymbol{z}}_j^o), \tag{23}$$

$$P(z^M|\hat{\boldsymbol{z}}_j^M, \hat{\boldsymbol{z}}_j^O) = \sum_z P(z^M|z)P(z|\hat{\boldsymbol{z}}_j^m, \hat{\boldsymbol{z}}_j^o). \tag{24}$$

These probabilities are also represented by finite and discrete parameters, which can be communicated by using an MP approach. These parameters are sent to an object concept model and a motion concept model, respectively. In the object and motion concept model, the latent variables that are assigned to the modality information $m \in \{v, a, h, p\}$ of concept $C \in \{O, M\}$ are determined by using Gibbs sampling.

$$z_{jmn}^C \sim P(z^C|\boldsymbol{W}^m, \boldsymbol{Z}_{-jmn})P(z^C|\hat{\boldsymbol{z}}_j^M, \hat{\boldsymbol{z}}_j^O), \tag{25}$$

where $\boldsymbol{W}^m$ represents all information of modality $m$, and $\boldsymbol{Z}_{-jn}^o$ and represents a set of latent variables, except for a latent variable assigned to the information of modality $m$ of the $j$-th observation. Whereas the latent variables are sampled from $P(z^C|\boldsymbol{W}^m, \boldsymbol{Z}_{-jmn})$ in the normal MLDA, they are sampled by using $P(z^C|\hat{\boldsymbol{z}}_j^M, \hat{\boldsymbol{z}}_j^O)$ along with it. Therefore, the all latent variables are learned in a complementary manner. From the sampled variables, the parameters of $P(z_j^o|\boldsymbol{w}_j^v, \boldsymbol{w}_j^a, \boldsymbol{w}_j^h)$ and $P(z_j^m|\boldsymbol{w}_j^m)$ are updated, and equations (21)-(25) are iterated until converged.

Fig. 7 shows the pseudo code of mMLDA, and the corresponding graphical model. The model shown in the left of Fig. 7 can be constructed by connecting the latent variables based on Serket. Although the part framed by the red rectangle is implemented in the experiment, it can be easily extended to the model shown in this figure.

### 4.2 Experimental Results

By using the model explained in Sec. 4.1, the object categories, motion categories and integrated categories that represent the relationships among them were formed from visual, auditory, haptic and motion information obtained by the robot. The information obtained by the robot is detailed in the Appendix. We compared it with an independent model where the object and motion categories are learned independently.

Fig. 8(a) shows a confusion matrix of classification by the model where the object and motion categories are learned independently, and the vertical and horizontal axis respectively represent the correct category index as well as the category index to which each object is classified. The accuracies were 98% and 72%. One can see that the motion categories can be formed by the independent model almost correctly, however, the object categories cannot be formed correctly compared to the motion categories. On the other hand, Fig. 8(b) shows the result by using mMLDA implemented based on Serket, and the categories are learned in a complementary manner. The classification accuracies were 100% and 94%. The motion that could not be classified correctly by the independent model was able to be classified correctly. Moreover, the accuracy of object classification received a 22% improvement thanks to the effect of the motion categories. In the independent model, the objects of category five (shampoos) were classified into category seven because of their visual similarity. On the other hand, in the mMLDA based on Serket, they were misclassified into category three (dressings) because the same motion (pouring) is acted upon these objects. Also, the rattles (category ten) were misclassified because the rattles (category ten) and soft toys (category nine) had a similar appearance and the same motion (throwing) were acted upon them. However, other objects were classified correctly, and this fact indicates that mutual learning is realized by Serket.

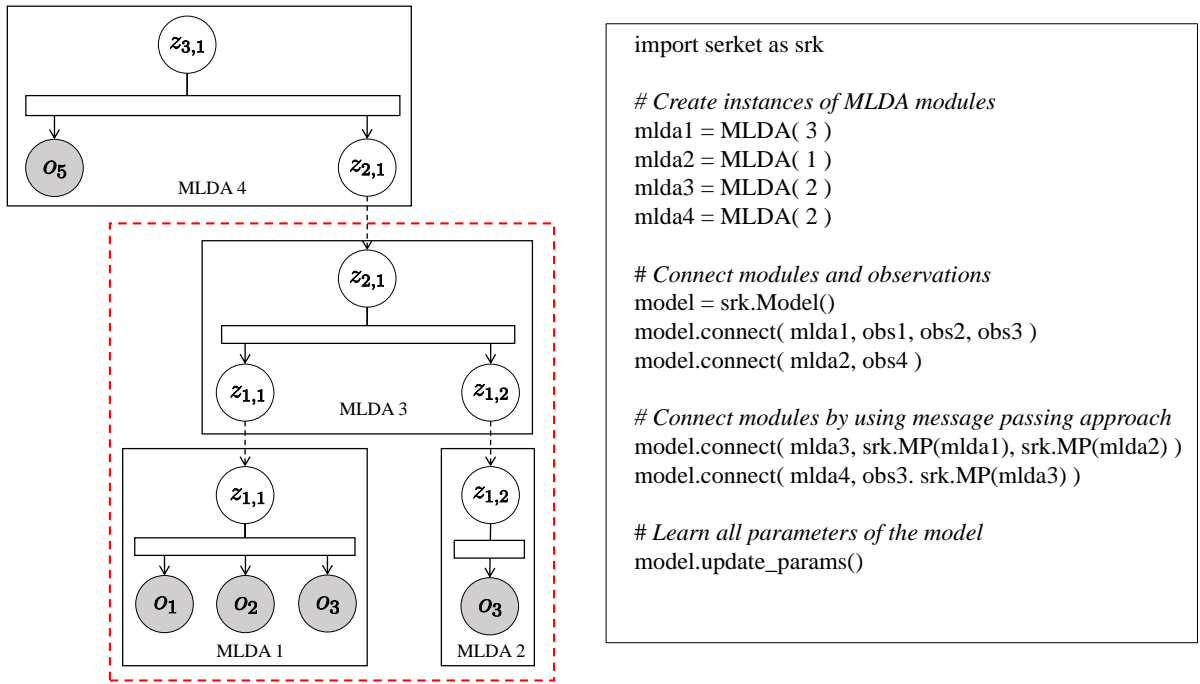

**Figure 7.** Pseudo code of mMLDA

Furthermore, we conducted an experiment to investigate the efficiency of the original mMLDA that are not divided into modules. As the results display in Fig. 8(c), the accuracies of the classification of objects and motions were respectively 100% and 94%, although misclassified objects are different from that of the Serket implementation of mMLDA because of sampling. One can see that mMLDA implemented based on Serket is comparable with the original mMLDA. In the original mMLDA, the structure of the model was fixed, and we derived the equations to estimate its parameters and implemented it. On the other hand, by using Serket, we can flexibly change the structure of the model as shown in Fig. 7 without deriving the equations for the parameter estimation.

## 5 EXAMPLE 2: MUTUAL LEARNING OF CONCEPT MODEL AND LANGUAGE MODEL

### 5.1 Implimentation Based on Serket

In (Nakamura et al., 2014; Nishihara et al., 2017), we proposed a model for the mutual learning of concepts and the language model shown in Fig. 4(b), and its parameters were approximately estimated by dividing the models into smaller parts. Its approximation for the parameter estimation is considered equivalent to the SIR approach of Serket. Here, we reconsider it based on Serket. The model shown in Fig. 4(b) is a model where the speech recognition part and the MLDA that represents the object concepts are connected, and can be divided as shown in Fig. 9. The MLDA makes it possible to form object categories by classifying the visual, auditory and haptic information obtained as shown in Fig.12. In addition, the words in the recognized strings of a user's utterances to teach object features are also classified in the model shown in Fig. 9. By this categorization of multimodal information and teaching utterances, the words and multimodal information are connected stochastically, and this enables the robot to infer the sensory information represented by the words. However, the robot cannot obtain the recognized strings directly, it can obtain only continuous speech. Therefore, in the model shown in Fig. 9, the words $s$ that are in the recognized strings are treated as latent variables and connected to the model for speech recognition. The parameter $\mathcal{L}$ of the language model is also a latent variable, and is learned from recognized strings of continuous speech $o$ by using NPYLM (Mochihashi et al., 2009). Furthermore, it is an important point of this model that MLDA and the speech recognition model are connected through the words $s$. This makes it possible to learn them in a complementary manner. That is, the speech is not only recognized based on the similarity of $o$ but is accurately recognized by utilizing the inferred words $s$ from the multimodal information perceived by the robot.

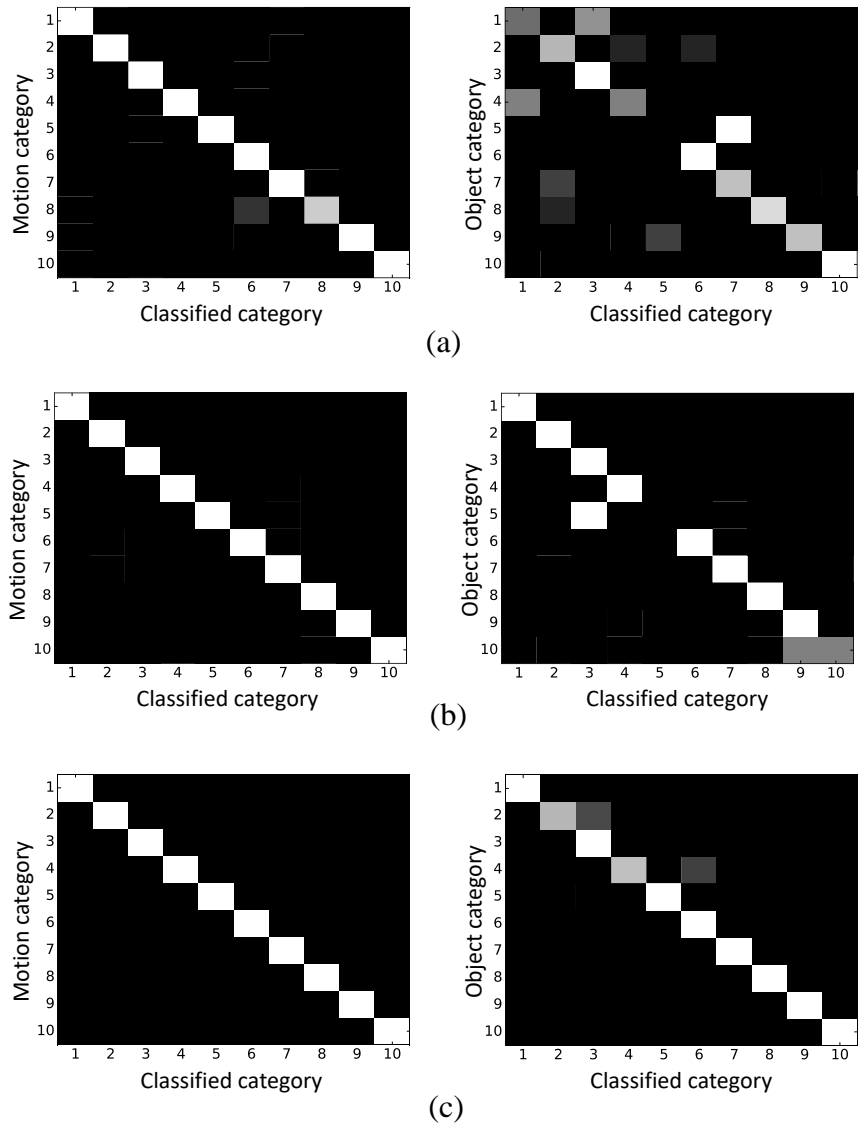

**Figure 8.** Classification result of motion and object by (a) independent model, (b) Serket implementation, and (c) original model. The classification accuracies motions and objects were (a) 98% and 72%, (b) 100% and 94%, and (c) 100% and 94%.

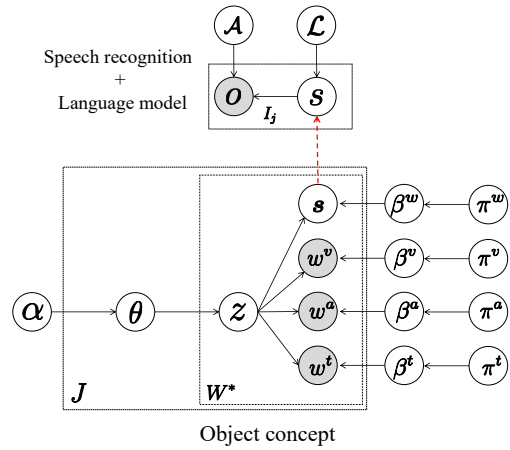

**Figure 9.** Mutual learning model of concepts and language model.

First, as the initial parameter of $\mathcal{L}$, we use the language model where all phonemes are generated with equal probabilities. The MP approach can be used if all teaching utterances $O$ are recognized by using a language model whose parameter is $\mathcal{L}$ and the probability $P(S|O, \mathcal{A}, L)$ that the word sequences $S$ are generated can be computed. However, it is actually difficult to compute the probabilities for all possible word segmentation patterns of all possible recognized strings. Therefore, we approximate this probability distribution by using an SIR approach. The $L$-best speech recognition results are utilized as samples because it is difficult to compute the probabilities for all possible recognized strings. $s_j^{(l)}$ represents the $l$-th recognized string of a teaching utterance given the $j$-th object. By applying NPYLM and segmenting them into words, the word sequences $S = \{s_j^{(l)} | 1 \leq l \leq L, 1 \leq j \leq J\}$ can be obtained.

$$S \sim P(S|S', \mathcal{L}). \tag{26}$$

These generated samples are sent to the MLDA module, and the samples that are likely to represent multimodal information are sampled among them based on the MLDA whose current parameter is $\Theta$:

$$\hat{s}_j \sim P(s_j^{(l)} | w_j^v, w_j^a, w_j^t, \Theta). \tag{27}$$

The selected samples $\hat{s}_j$ are considered as words that can represent multimodal information. Then, the parameters of MLDA are updated by using a set of these words $\hat{S} = \{\hat{s}_j | 1 \leq j \leq J\}$ and a set of multimodal information $W^v, W^a, W^t$ by utilizing Gibbs sampling.

$$\Theta = \arg\max P(\hat{S}, W^v, W^a, W^t | \Theta). \tag{28}$$

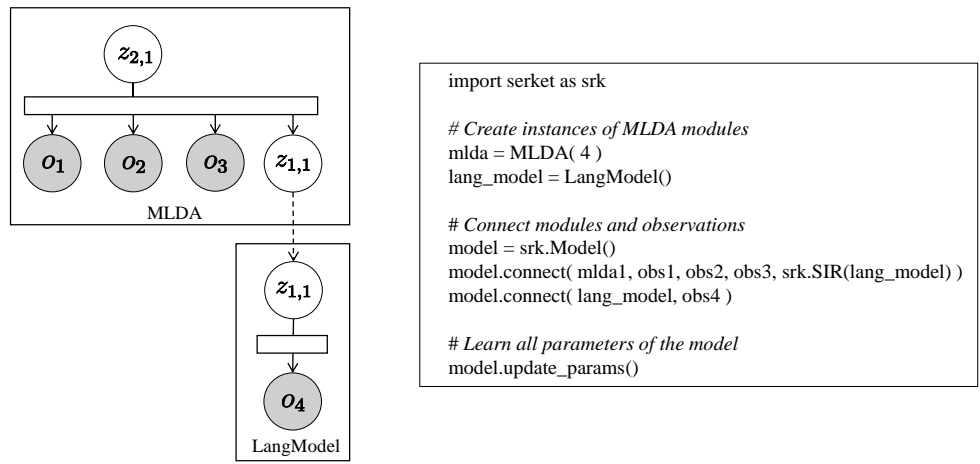

**Figure 10.** Pseudo code of mutual learning of concept model and language model.

Moreover, $\hat{S}$ is sent to the speech recognition model, and the parameter $\mathcal{L}$ of the language model is updated.

$$\mathcal{L} = \arg\max P(\hat{S}|\hat{S}', \mathcal{L}), \tag{29}$$

where $\hat{S}'$ denotes strings obtained by connecting words in $\hat{S}$. The parameters of the whole model can be optimized by iterating through the above process: the sampling words and using Eq. (26), the resampling words using Eq. (27), and the updating parameters using Eq. (28) and Eq. (29).

Fig. 10 displays the pseudo code and the corresponding graphical model. In this model, one of modules is MLDA and has three observations and one shared latent variable that is connected to the speech recognition module. $o_1$, $o_2$ and $o_3$ represent multimodal information obtained by the sensors on the robot, and $o_4$, which is an observation of the speech recognition model, represents the utterances given from the human user. Although the parameter estimation of the original model proposed in (Nakamura et al., 2014; Nishihara et al., 2017) is very complicated, it can be briefly described by connecting the modules based on Serket.

## 5.2 Experimental Results

We conducted an experiment where the concepts are formed using the aforementioned model to demonstrate the validity of Serket. We compared the following three methods.

(a) A method where speech recognition results $S'_0$ of teaching utterances with maximum likelihoods were segmented into words by applied NPYLM, and the words obtained were used for concept formation.
(b) A method where the concepts and language model are learned by a mutual learning model implemented based on Serket. (Proposed method)
(c) A method where the concepts and language model are learned by a mutual learning model implemented without Serket proposed in (Nakamura et al., 2014). (Original method)

**Table 1.** Accuracies of speech recognition, segmentation and object classification.

| Methods | (i) Speech recognition | (ii) Segmentation | | | (iii) Object classification |
|---|---|---|---|---|---|
| | | Precision | Rcall | F-measure | |
| (a) w/o mutual learning | 0.64 | 0.50 | 0.68 | 0.58 | 0.80 |
| (b) Serket implementation | 0.74 | 0.91 | 0.59 | 0.72 | 0.94 |
| (c) Original model | 0.77 | 0.95 | 0.59 | 0.73 | 0.94 |

**Table 2.** Evaluation of segmentation

| | (a) | (b) | (c) | (d) | (e) | (f) | (g) |
|---|---|---|---|---|---|---|---|
| Correct segmentation: | A | / | B | | C | / | D |
| Estimated segmentation: | A | / | A | / | C | | D |
| Evaluation: | TN | TP | TN | FP | TN | FN | TN |

In method (a), the following equation was used instead of Eq. (26), and the parameter $\mathcal{L}$ of language model is not updated:

$$\boldsymbol{S}_0 \sim P(\boldsymbol{S}|\boldsymbol{S}'_0, \mathcal{L}). \tag{30}$$

Alternatively, method (b) is implemented by Serket, and the concepts and language model are learned mutually through the shared latent variable $s$.

Tbl. 1(i) shows the speech recognition accuracies of each method. In method (a), the language model was not updated, and, therefore, the accuracy is equal to phoneme recognition. Instead, in method (b), the accuracy is higher than that of method (a) by updating the language model from the words sampled by MLDA.

Tbl. 1(ii) shows the accuracies of word segmentation. Segmentation points are evaluated as shown in Tbl. 2 by applying DP (dynamic programming) matching to find the correspondence between the correct and estimated segmentation. This table shows a case where the correct segmentation of a correctly recognized string "ABCD" is "A/BC/D" and the recognized string "AACD" is segmented into "A/A/CD". ("/" represents the cut points between each word. ) The points that are correctly estimated in (Tbl. 2(b)) as cut points were evaluated as TP (true positive), and those that are not correctly estimated (Tbl. 2(d)) were evaluated as FP. Similarly, the points that are erroneously estimated as not cut points (Tbl. 2(f)) were evaluated as FN (false negative). From the evaluation of the cut points, the precision, recall and F-measure are computed as follows.

$$P = \frac{N_{TP}}{N_{TP} + N_{FP}}, \tag{31}$$

$$R = \frac{N_{TP}}{N_{TP} + N_{FN}}, \tag{32}$$

$$F = \frac{2RP}{R + P}, \tag{33}$$

where $N_{TP}$, $N_{FP}$ and $N_{FN}$ denote the number of points that are evaluated as TP, FP and FN, respectively. Comparing the precision of methods (a) and (b) in Tbl. 1(ii), one can see that it increases according to Serket. This is because more correct words can be selected among the samples generated by the speech recognition module. Alternatively, the recall of method (b) decreases, and this is because some functional words (e.g. "is" and "of") are connected with other words such as "bottleof." However, the precision of method (b) is higher, and, moreover, the F-measure of it is higher than 0.11. Therefore, method (b), which is implemented based on Serket, outperformed method (a). Furthermore, Tbl. 1(iii) displays the object classification accuracy, and one can observe that the accuracy of method (b) is higher because the speech can be recognized more correctly.

Moreover, the Serket implementation (method (b)) was comparable to the original implementation (method (c)). Thus, the learning of the object concepts and language model which were presented in (Nakamura et al., 2014; Nishihara et al., 2017) is realized by Serket.

# 6 CONCLUSION

In this paper, we proposed a novel architecture where the cognitive model can be constructed by connecting modules, each of which maintains programmatic independence. Two approaches are used for connecting them. One is the MP approach where the parameters of the distribution are of a finite dimension and are communicated between the modules. If the parameters of the distribution are of an infinite dimension or a complex structure, the SIR approach is utilized to approximate them. In the experiment, we demonstrated two implementations based on Serket and their efficiency. The experimental results demonstrated that the implementations are comparable with the original models.

In this paper, we focus on the connected generative probabilistic models. However, we believe that models that can be connected by Serket are not limited to generative probabilistic models. Neural networks or other methods can be one of modules of Serket, and we are planning to connect them. Furthermore, we believe that large scale cognitive models can be constructed by connecting various types of modules, each of which represents a particular brain function. In so doing, we will realize our goal of artificial general intelligence. Serket can also contribute to developmental robotics (Asada et al., 2009; Cangelosi et al., 2015) where the human developmental mechanism is understood via a constructive approach. We believe robots that can learn capabilities ranging from motor skills to language, and this can be developed using Serket, as it makes it possible to understand humans.

# APPENDIX 1: FUNDAMENTAL MODELS

In this section, we detailed the multimodal LDA and learning of the language model, which are used for two implementations in this paper.

**Multimodal LDA**

We used multimodal latent Dirichlet allocation (MLDA) as one of modules. MLDA is an extension of LDA (Blei et al., 2003), which is proposed for document classification, to classify multimodal information obtained by the robot's sensors. In this model, it is assumed that the multimodal information $w_1, w_2, w_3$ is generated by following generative process:

- Category ration is determined:

$$\theta \sim P(\theta|\beta). \tag{34}$$

- Following process is iterated $N_m$ times for $m \in \{1, 2, 3\}$:

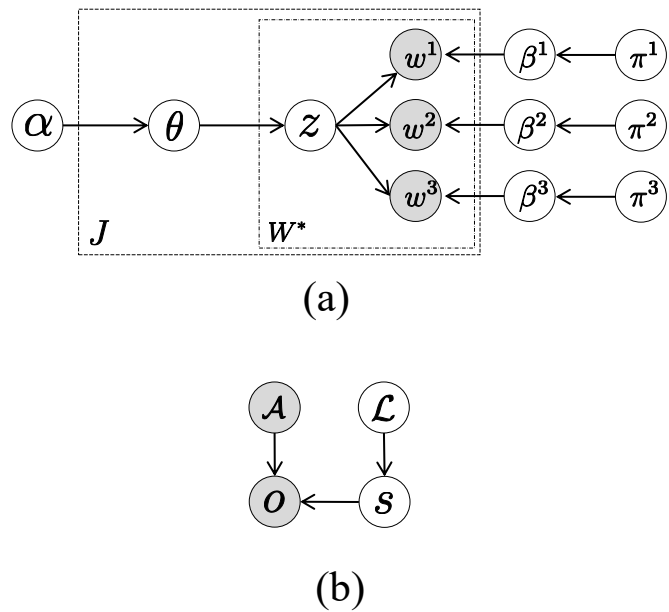

**Figure 11.** (a) Model for concept formation, and (b) model for learning of the language model.

1. A category is selected:

$$z \sim P(z|\theta). \tag{35}$$

2. Information of category $z$ is generated:

$$w_m \sim P(w|\phi_{mz}). \tag{36}$$

MLDA stochastically models the generative process of multimodal information, and multimodal information $w_m$ is assumed to be sampled from the distribution $P(w_m|\phi_{mz})$, such that the information $w_m$ of the category $z$ of modality $m$ is generated. Fig. 11(a) is a graphical model of MLDA, and depicts this generative process of multimodal information. As well as LDA, the categories can be learned in an unsupervised manner using Gibbs sampling where category $z$ is sampled and the model parameters $\theta, \phi_m$ are estimated. MLDA is the most fundamental model for multimodal categorization, and it can be extended to the multimodal hierarchical Dirichlet process (Nakamura et al., 2011), which makes it possible to estimate the number of categories, as well as an infinite mixture of models (Nakamura et al., 2015), which makes it possible to estimate the model structure.

In terms of multimodal information, we used visual, auditory and haptic information, which we explain later. In the previous study (Nakamura et al., 2007), we indicated that more human-like categories can be formed by a classification of multimodal information.

**Learning of language model**

The robot can form object concepts using MLDA, and acquire word meaning by connecting formed concepts and words that are taught through an interaction with others. To obtain the words, the robots are required to recognize speech and extract words from it. In order to do that, a language model is required and can be learned in an unsupervised manner using the model shown in Fig. 11(b). The variable $o$ represents the given human speech, and this is recognized and converted into a sequence of words $s$ using parameters of the acoustic model $\mathcal{A}$ and language model $\mathcal{L}$. Here, we consider that $\mathcal{A}$ is already known and $\mathcal{L}$ is learned. In the initial learning phase, the parameters of the language model are unknown, and we set it to a uniform distribution where all phonemes are equally generated. First, the parameter of the language model $\mathcal{L}$ can be estimated by dividing the recognized strings into words using a nested Pitman-Yor language model (NPYLM) (Mochihashi et al., 2009), which is a method for unsupervised morphological analysis. This word segmentation is realized via an estimation parameter $\mathcal{L}$ that maximizes the probability that the word sequence $\boldsymbol{S} = \{w_1^w, w_2^w, \cdots\}$ of recognized strings $\boldsymbol{S}'$ is generated.

$$\mathcal{L}, \boldsymbol{S} = \underset{\mathcal{L}, \boldsymbol{S}}{\operatorname{argmax}} P(\boldsymbol{S}|\boldsymbol{S}'\mathcal{L}). \tag{37}$$

The learned language model can enable the robot to recognize speech accurately.

### 6.0.1 Hierarchical Pitman-Yor Language Model

The hierarchical Pitman-Yor language model (HPYLM) is an n-gram language model in which the hierarchical Pitman-Yor process is used. In the HPYLM, the probability that a word $w$ appears after a context $h$ is computed as follows:

$$p(w|h) = \frac{c(w|h) - d \cdot t_{hw}}{\theta + \sum_w c(w|h)} + \frac{\theta + d \cdot \sum_w t_{hw}}{\theta + \sum_w c(w|h)} p(w|h'), \tag{38}$$

where $h'$ represents an $(n-1)$-gram context, $p(w|h')$ is the probability that the word $w$ appears after the context that is one shorter than $h$, and, therefore, these probabilities are can be computed recursively. Also, $c(w|h)$ represents the number of occurrence of $w$, and $t_{hw}$ represents the number of occurrences of $w$ in the context $h$. $d$ and $\theta$ are the hyper parameters of the Pitman-Yor process.

### 6.0.2 Nested Pitman-Yor Language Model

In the HPYLM mentioned in the previous section, $p(w|h')$ in Eq. (38) can be set as the reciprocal of the number of vocabulary in the case of unigram. However, we assumed that the vocabulary is not predefined, and it is difficult to compute it because all possible substrings in the recognized strings can be words. In order to solve this problem, the character HPYLM is used as the base measure of the word unigram. This model is called the Nested Pitman-Yor Language Model (NPYLM) since the character HPYLM is embedded in the word HPYLM. By utilizing the blocked Gibbs sampler and dynamic programming, NPYLM can divide strings into words efficiently.

## APPENDIX 2: MULTIMODAL OBJECT DATASET

Fig. 12(d) displays the objects used in the experiments. The robot obtained the multimodal information from these objects by observing, grasping and shaking them.

> **Visual information** $w^v$  A CCD camera and depth sensor are mounted on the arm of the robot (Fig. 12(a)), and the images captured by observing the objects are utilized for visual information. A Dense Scale Invariant Feature Transform (DSIFT) (Vedaldi and Fulkerson, 2010) is computed from each image, and each feature vector is quantized using 500 representative vectors, and converted into a 500-dimensional histogram.

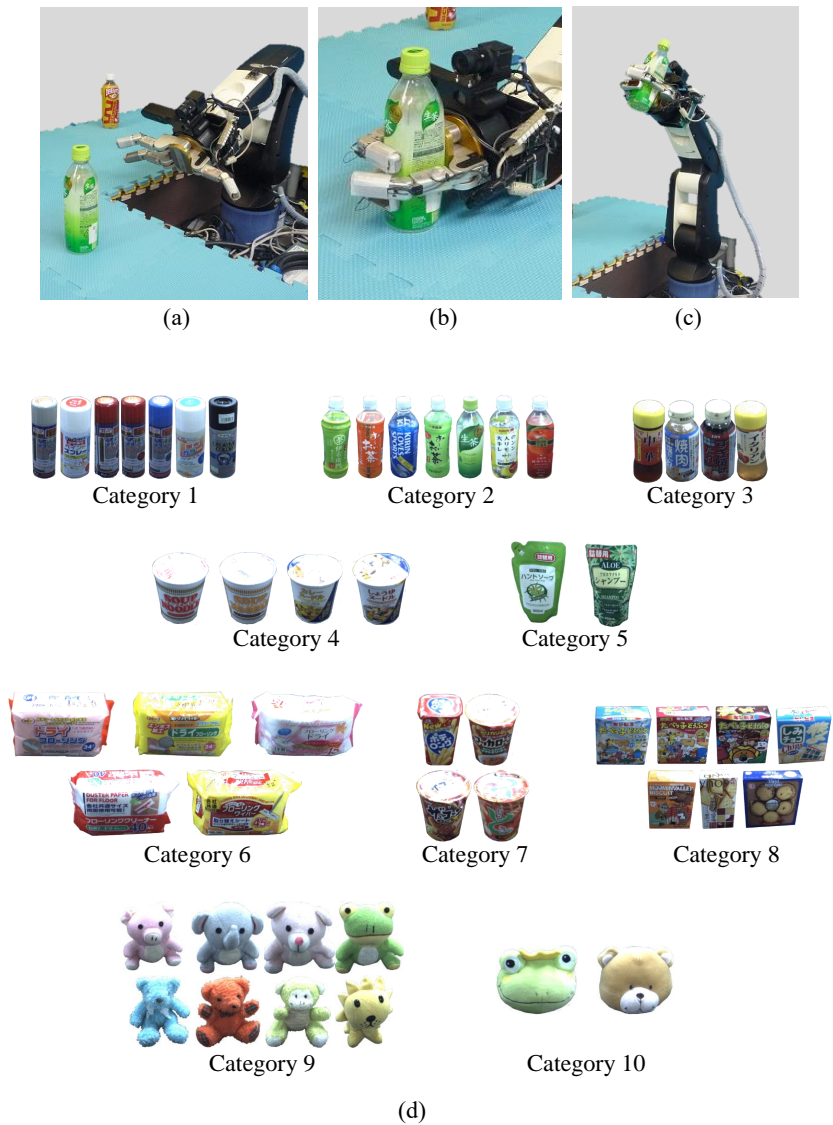

**Figure 12.** Obtaining (a) visual information, (b) haptic information and (c) auditory information. (d) 50 objects used in the experiments.

**Haptic information** $w^t$

Haptic information is obtained using a Barrett hand mounted on the arm, and a tactile array sensor is mounted on the hand (Fig. 12(b)). The robot grasps the objects and obtains a time-series of sensor values. The sensor values are approximated by a sigmoid function, the parameters of which are used as feature vectors (Araki et al., 2011). Finally, these feature vectors are quantized and converted into a 15-dimensional histogram.

**Auditory information** $w^a$

A microphone is mounted on the robot's hand, and the sound is captured by shaking the objects (Fig. 12(c)). The sound is divided into frames, and a 13-dimensional MFCC (Mel-Frequency Cepstrum Coefficient) is computed from each frame. Therefore, the sound is converted into a 13-dimensional feature vector. As well as the other information, these feature vectors are quantized and converted into a 50-dimensional histogram.

**Table 3.** Motions that are carried out against objects.

| motion | object | motion | object |
|---|---|---|---|
| pour (1) | dressing | wipe (5) | flooring cleaner |
| | shampoo | spray (6) | spray can |
| shake (2) | spray can | look (7) | soft toy |
| | plastic bottle | put (8) | snack |
| | dressing | | cup noodle |
| drink (3) | plastic bottle | throw (9) | soft toy |
| eat (4) | cup noodles | | rattle |
| | snack | pick up (10) | cookie |

**Motion information** $w^p$

Motion information is computed from the joint angles captured by Microsoft Kinect. The sequence of 11 joints angles are captured. We assumed that each motion can be segmented based on the identity of the manipulated object, and, therefore, the sequence can be considered from the beginning to the end of the manipulation of each object as one motion. Tbl. 3 displays the motions carried out against each object. The 11 joint angles are treated as 11 dimensional feature vectors, these feature vectors are quantized and converted into a 70-dimensional histogram. This histogram is a bag of features representation of the motions, the efficiency of which is shown in (Mangin and Oudeyer, 2012).

**Teaching utterances** $o$

The speech that a human user provides to teach object features is used as the teaching utterances. Each speech corresponds to each object based on the object identities. Therefore, the speech uttered during a robot's observing, grasping and shaking is assumed to represent its object feature.

In the experiment, the multimodal information was obtained through the following procedure. First, the user put an object in front of the robot. After detecting the object, the robot observed, grasped and shook it to obtain the multimodal information. Simultaneously, the user teaches the object features by speech. We instructed the user to teach the object features, and did not impose any limitation for their expression.

## ACKNOWLEDGEMENT

This work was supported by JST CREST Grant Number JPMJCR15E3.